%% file: main.tex
\documentclass[10pt,twocolumn,letterpaper]{article}

\usepackage[pagenumbers]{cvpr}
\usepackage{multirow}
\usepackage{booktabs} 
\usepackage{siunitx} 
\input{preamble}
\definecolor{cvprblue}{rgb}{0.21,0.49,0.74}
\usepackage[pagebackref,breaklinks,colorlinks,allcolors=cvprblue]{hyperref}

\title{PortraitDirector: A Hierarchical Disentanglement Framework for Controllable and Real-time Facial Reenactment}

\author{
    Chaonan Ji \quad Jinwei Qi \quad Sheng Xu \quad Peng Zhang \quad Bang Zhang \\
    Tongyi Lab
}

\begin{document}

\twocolumn[{
    \renewcommand\twocolumn[1][]{#1}
    \maketitle
    \vspace*{-2.9em}
    \begin{center}
        \captionsetup{type=figure}
        \includegraphics[width=1.0\linewidth]{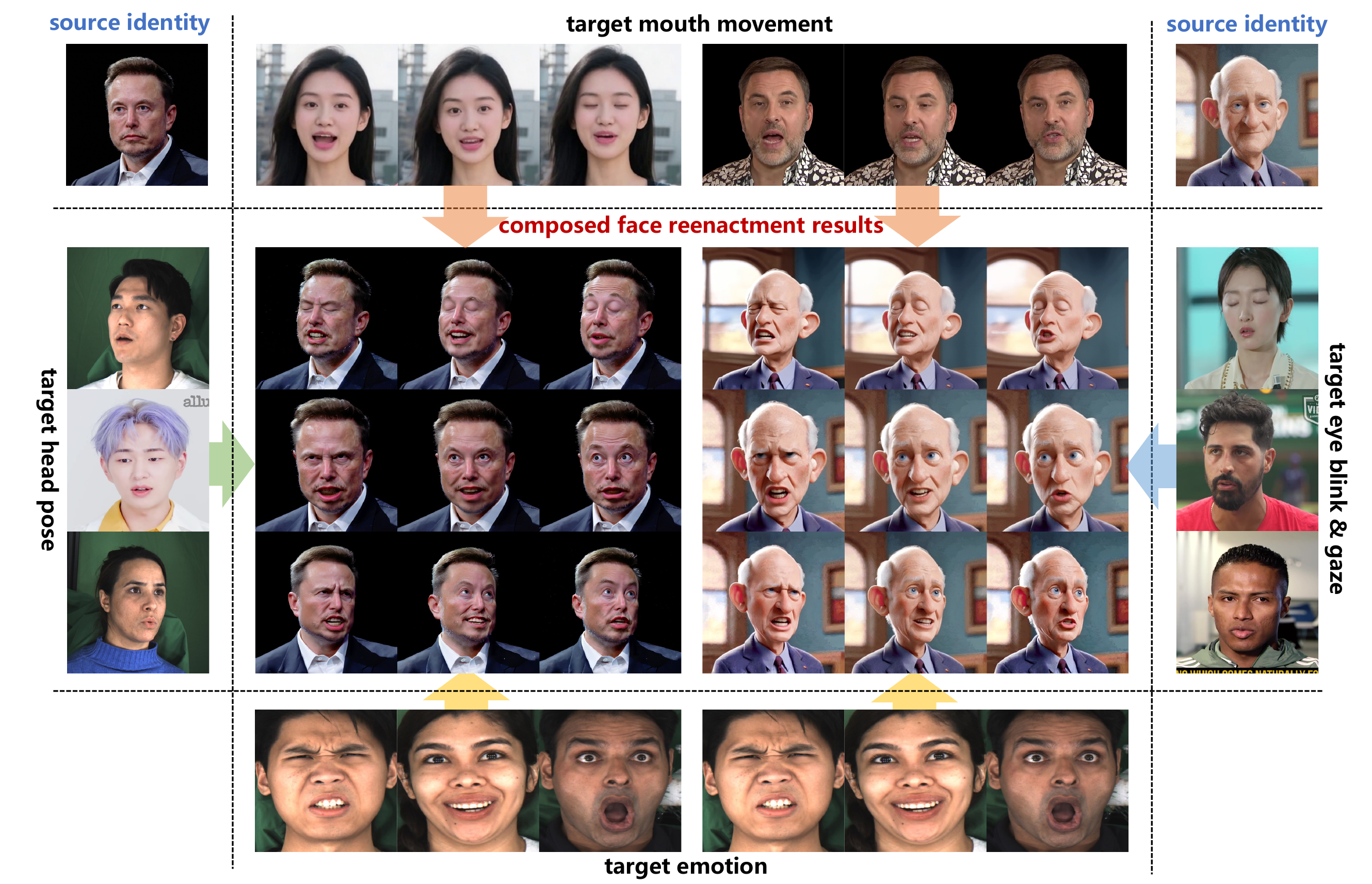} \vspace{-0.8em}
        \captionof{figure}{
            Our model generates face reenactment results (middle) from a reference image with disentangled control over mouth motion, head pose, eye motion, and emotion. Each motion component can be independently controlled by a different driving video. 
        }
        \label{fig:overview}
    \end{center}
}]

\maketitle
\input{sec/0_abstract}    
\input{sec/1_intro}

\input{sec/2_related}
\input{sec/3_method}

\input{sec/4_experiment}

\input{sec/5_conclusion}
{
    \small
    \bibliographystyle{ieeenat_fullname}
    \bibliography{main}
}


\end{document}

%% file: sec/0_abstract.tex
\begin{abstract}

Existing facial reenactment methods struggle with a trade-off between expressiveness and fine-grained controllability. Holistic facial reenactment models often sacrifice granular control for expressiveness, while methods designed for control may struggle with fidelity and robust disentanglement. Instead of treating facial motion as a monolithic signal, we explore an alternative compositional perspective. In this paper, we introduce PortraitDirector, a novel framework that formulates face reenactment as a hierarchical composition task, achieving high-fidelity and controllable results. 
We employ a \textbf{Hierarchical Motion Disentanglement and Composition} strategy, deconstructing facial motion into a Spatial Layer for physical movements and a Semantic Layer for emotional content. The Spatial Layer comprises: (i) global head pose, managed via a dedicated representation and injection pathway; (ii) spatially separated local facial expressions, distilled from cropped facial regions and purged of emotional cues via \textbf{Emotion-Filtering Module} leveraging an information bottleneck. The Semantic Layer contains a derived global emotion. 
The disentangled components are then recomposed into an expressive motion latent.
Furthermore, we engineer the framework for real-time performance through a suite of optimizations, including diffusion distillation, causal attention and VAE acceleration. PortraitDirector achieves streaming, high-fidelity, controllable $512 \times 512$ face reenactment at 20\,FPS with a end-to-end 800\,ms latency on a single 5090 GPU.
\end{abstract}

%% file: sec/1_intro.tex
\section{Introduction}
\label{sec:intro}
Facial reenactment animates a source identity with motions from a driving video, aiming for photorealistic, identity-preserving results. Beyond this, a key goal is offering fine-grained control over individual components like expression and head pose, which is crucial for applications in filmmaking, virtual avatars, and interactive entertainment.

While existing methods \cite{wav2lip,obama,styleSync,SyncTalkFace,SongWQHL22,SongZLWQ19,ThiesETTN20,ChenLMDX18,ZhangLGH24,GuoCLLBZ21,JiZWWWXC22,LiangPGZHHHLD022,SunZ0K21,Wu0WDDD21,WangM021,DPE,WangZSZ0YL23,TanJP24,LiuC0DC0024,metaportrait,EMOPortraits,VASA,float,liveportrait,lia,megaportrait,sadtalker,xportrait2,aniportrait,fantasy,hunyuan,edtalk,edtalk++,eat,swaphead,pctalk,fixtalk} have made notable strides, they face a trade-off between expressiveness and controllability. On one side of the spectrum, holistic, end-to-end models \cite{xportrait2,fantasy,aniportrait,EMOPortraits} excel in generating highly expressive and photorealistic results by learning a single, entangled motion representation. However, this monolithic approach inherently sacrifices fine-grained control ability, thereby inherently limiting the ability to independently manipulate specific aspects of the animation. 
Conversely, other methods prioritize controllability. Approaches based on explicit geometric proxies \cite{WangZSZ0YL23,aniportrait,emoji} are inherently constrained by the predefined capacity of the proxies themselves. More recent works, such as EDTalk \cite{edtalk} and PD-FGC \cite{pdfgc}, have pioneered efforts to disentangle the latent motion space for more granular control. These methods typically enforce separation through specialized training objectives and data augmentation strategies. While demonstrating the potential of disentangled control, this reliance on indirect supervision can make it challenging to achieve consistently robust separation and fine-grained control. 

We posit that this persistent challenge stems in part from a common tendency to model facial motion as a monolithic signal. Our key insight is to treat the facial motion not a single, entangled signal, but rather a multi-layered composition of different motion components, where each operates at distinct spatial, semantic and temporal scales. This hierarchical perspective is pivotal, unlocking fine-grained control by formulating reenactment from a `driving' to a `composition' task. To this end, we introduce PortraitDirector, a hierarchical, coarse-to-fine real-time framework that achieves high-fidelity and controllable generation by strategically composing these distinct motion components.

Specifically, We first disentangle the input motion by routing its components to specialized layers based on the properties of its constituent components, which are later recombined to reconstruct the complete motion. The Spatial Layer isolates head pose and crops facial Regions-of-Interest (ROIs). First, to ensure pose-expression disentanglement, we decouple their representations and injection pathways: head pose is encoded into a dedicated latent space and injected independently of the facial expression stream. Second, cropping dedicated ROIs for each facial component (eg,.eyes), enforces their disentanglement and prevents motion leakage. The Semantic Layer derives a global emotion via temporal pooling of motion latents. Finally, the Composite Layer reassembles these components into a unified control signal. This architectural choice of initial decomposition is crucial for preventing feature entanglement and enabling precise, regional facial control. 

While the aforementioned strategy yields promising decomposition results, 
we observe that the emotional expression from the driving image heavily biases the final output. Specifically, the driving video's inherent emotion dominates the synthesis, causing the output to ignore or resist the target emotion.
To address this issue, we introduce an Emotion-Filtering Module to disentangle the emotional expression from the facial movements. Specifically, we first employs an autoencoder-based information bottleneck to distill local motion component into an emotion-agnostic basis and then utilizes a cross-attention mechanism to re-integrate this neutral basis with the target emotional expression. This ensures expressive control by breaking the inherent entanglement between facial motion and emotion.

Furthermore, we place a strong emphasis on computational efficiency. Our framework is engineered for real-time performance, achieved through a suite of key optimizations: Distribution Matching Distillation to reduce sampling steps, a redesigned causal attention mechanism for streaming, and VAE decoder acceleration. This allows the pipeline to achieve 20\,FPS a low latency of 800\,ms at $512 \times 512$ resolution on a single 5090 GPU, making our high-fidelity, controllable reenactment practical for live applications.

To conclude, our main contributions are the following:

\begin{itemize}

\item A Hierarchical Motion Disentanglement and Composition Framework built on the principle of pose-expression disentanglement and facial motion disentanglement, enabling the strategic composition of distinct motion components to achieve high-fidelity, controllable results.
\item An Emotion-Filtering Module that leverages the Information Bottleneck principle to decouple facial movements from emotional expressions.
\item Achieving Real-time streaming performance at 20\,FPS for $512 \times 512$ resolution, with a low latency of 800\,ms.

\end{itemize}

%% file: sec/2_related.tex
\section{Related Work}
\label{sec:relatedwork}

\subsection{Facial Reenactment}
Facial reenactment has been an active and prominent research area, with research largely split between two main philosophies for motion representation: structured geometric signals versus learned latent codes.

Previous methods \cite{WangZSZ0YL23,aniportrait,metaportrait,meshportrait,emoji,DBLP:conf/cvpr/TewariEBBSPZT20,dvp,face2face} leverage explicit geometric proxy, such as 3D Morphable Model (3DMM) \cite{WangZSZ0YL23, faceverse, mediapipe} or facial landmarks \cite{aniportrait, magicpose,emoji}. This provides a degree of interpretable control, as exemplified by AniPortrait \cite{aniportrait} and StyleAvatar \cite{WangZSZ0YL23}, which use 3DMM-derived landmarks for high-quality cross-identity generation. While these methods exhibit some motion decomposition capabilities, they are ultimately constrained by the representational bottleneck of their geometric proxies. This bottleneck prevents a clean separation of motion components and, crucially, precludes the isolation of emotion from the underlying physical articulation.

To enable more expressive facial control, some methods \cite{EMOPortraits,VASA,float,liveportrait,lia,megaportrait,sadtalker,xportrait2,fantasy,hunyuan,edtalk,edtalk++,animatex,DBLP:conf/iccv/BounareliTAPT23,oorloff2023oneshotfacevideoreenactment} adopts learned latent representations to capture subtle expression.
Face-vid2vid \cite{WangM021}, SadTalker \cite{sadtalker}, and LivePortrait \cite{liveportrait} utilize implicit keypoints derived from images as a compact and powerful signal to guide the animation process.  Methods such as MegaPortrait \cite{megaportrait}, EMOPortrait \cite{EMOPortraits}, VASA \cite{VASA}, along with LIA \cite{lia} try to forgoes geometric proxies entirely and instead learn to encode motion directly into a low-dimensional vector. A new wave of methods, including XPortrait \cite{xportrait}, Xpotrait2 \cite{xportrait2} FantasyPortrait \cite{fantasy}, and HunyuanPortrait \cite{hunyuan}, incorporates diffusion models as the core generator and achieve superior synthesis capabilities. 
The holistic treatment of motion inherently entangles facial components.
We formulate reenactment as a compositional task, disentangling the signal into distinct parts and recomposing them to achieve precise, regional animation.

\subsection{Facial Motion Disentanglement}
The inherently correlated nature of facial dynamics, where head pose, mouth shape, and expression move in a holistic manner, poses a significant challenge for disentangled control. Prior works \cite{PC-AVS,TH-PAD,xportrait2,aniportrait,emoji,DPE,pdfgc,edtalk,eat,JiZWWWXC22,DBLP:conf/aaai/ZengPWZL20,DBLP:conf/cvpr/TranZHTHL24,DBLP:conf/cvpr/RochowSB24,DBLP:journals/corr/abs-2505-15313,DBLP:journals/tog/KimEZSBRT19,pumarola2020ganimation,cao2021unifacegan,papantoniou2022neural,wu2023poce,pernuvs2023maskfacegan} have attempted to isolate specific facial subspaces. PC-AVS \cite{PC-AVS} and TH-PAD \cite{TH-PAD} used contrastive learning to separate the mouth region.
DPE \cite{DPE} employs a bidirectional cyclic training strategy to separate pose and expression. PD-FGC \cite{pdfgc} pioneered with a coarse-to-fine latent decomposition, and EDTalk \cite{edtalk} further refined it by enforcing motion orthogonality.
However, the success of these methods is critically contingent upon carefully engineered data augmentations and auxiliary training objectives. XPortrait2 \cite{xportrait2} disentangles head pose and facial expression from the detected facial bounding box. Similarly, AniPortrait \cite{aniportrait} and Follow-Your-Emoji \cite{emoji} achieve a decoupling of pose and motion by using facial templates. They offer only partial and entangled control over facial attributes. To overcome these limitations, we design a Hierarchical Composition Framework empowered by a novel Emotion-Filtering Module, achieving a high-fidelity decomposition of facial dynamics and allowing for a clean and controllable separation of its core components.

\subsection{Diffusion Model Acceleration}
Reducing sampling steps is a key strategy for accelerating diffusion models. While early distillation methods like Latent Consistency Models (LCM)~\cite{LCM} achieve single-step inference, they often sacrifice quality and training stability; DMD~\cite{dmd} improves upon this by using a noise distribution loss to better align the student and teacher sampling trajectories.
Efficient sequential generation architectures, like those using KV caching~\cite{causvid}, suffer from a train-test discrepancy due to error accumulation. Self-Forcing~\cite{selfforcing} addresses this by training on the model's own outputs to align the train and inference distributions, while TalkingMachines~\cite{talkingmachines} further optimizes efficiency by re-engineering the attention mask.
We achieve real-time performance by enhancing Self-Forcing with: an efficient attention pattern and an accelerated VAE decoder for faster synthesis.

%% file: sec/3_method.tex
\section{Method}
\begin{figure*}[htb]
    \centering
    \includegraphics[width=\textwidth]{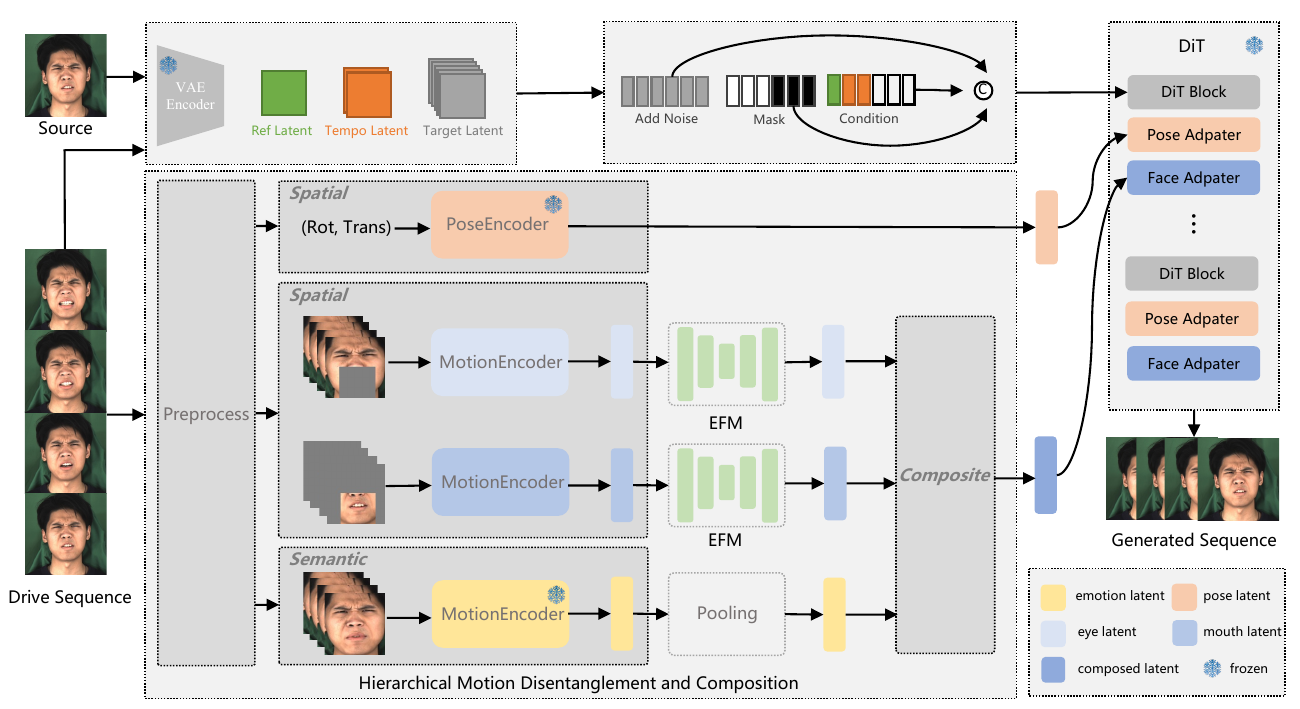}
    \caption{The overall framework. The driving sequence is first processed to extract pose latent and initial motion latents for different facial regions (e.g., eyes, mouth). The Emotion-Filtering Module (EFM) then removes inherent emotional cues from these regional latents, while a parallel branch extracts a global emotion latent via pooling. These refined regional latents are composed into a holistic motion latent, which, along with the pose latent, is injected into the DiT backbone via cross-attention. 
    Note that the weights of the DiT and the  MotionEncoder for emotion are kept frozen during this stage.
    }
    \label{fig:pipeline}
\end{figure*}

In this section, we first introduce the overall architecture of PortraitDirector framework in Sec.\ref{method:framework}. Then we elaborate on Hierarchical Motion Disentanglement and Composition in Sec.\ref{method:composite} and the Emotion-Filtering Module in Sec.\ref{method:EFM}. The Real-time Streaming Generation is detailed in Sec.\ref{method:fast}. An overview of the entire pipeline is provided in Fig.\ref{fig:pipeline}. 

\subsection{Overall Framework}
\label{method:framework}
Our model is designed based on the Wan-I2V \cite{wan2025wan} architecture, which we extend to support controllable facial reenactment. Following the methodology of the Wan-Animate \cite{wananimate}, we pretrain a foundational facial reenactment model, which consists of several components: 1) \textbf{MotionEncoder:} extracts motion latent $l_{face}$ from the cropped facial image 2) \textbf{PoseEncoder:} maps explicit head pose parameters $(Rot,Trans)$ into pose latent $l_{pose}$ 3) \textbf{Diffusion Transformer (DiT):} takes a noisy latent and the motion signals as conditioning to predict the denoised latent 4) \textbf{FaceAdapter and PoseAdapter:} are responsible for injecting the face and pose latent into the DiT through a series of dedicated cross-attention layers. The training objective is the standard diffusion denoising loss, but with an increased weight applied to the facial region:

\begin{equation}
    \begin{aligned}
        \mathcal{L} = \underset{t,x_{0},x_{1}, c}{\mathbb{E}}\left [ (1+\lambda M)\odot\left \| u\left ( x_{t},c,t;\theta  \right ) -v_{t}  \right \|\right ] 
    \end{aligned}
\end{equation}
where $x_{0}\sim \mathcal{N}\left ( 0,I \right )$, $x_{1}$ is the encoded image latent and $x_{t}=tx_{1}+(1-t)x_{0}$. $v_{t}=x_{1}-x_{0}$ is the ground truth velocity, $t$ is timestep and $c$ is the condition embedding. $M$ represents the binary mask for the facial region and $u\left ( x_{t},c,t;\theta  \right )$ denotes the predicted velocity. $\lambda$ is a hyperparameter and we empirically set its value to 50.

Leveraging the pretrained facial reenactment model, we introduce the 5) \textbf{Hierarchical Motion Disentanglement and Composition} module that disentangles facial motion into distinct spatial and emotional components and allows for targeted manipulation and faithful recomposition. This module serves as a drop-in replacement for the original MotionEncoder, mapping the driving image to a corresponding motion latent. During training, the pretrained DiT backbone is frozen, and we solely optimize the new module.

In addition to the denoising loss, we introduce a constraint on the composed latent to preserve expressiveness. We enforce that the composed latent should align with the latent produced by the original, pretrained MotionEncoder, which ensures the fidelity of the recomposed signal $l_{pred}$ relative to the original, holistic representation:
\begin{equation}
    \begin{aligned}
    \mathcal{L}_{latent} = {\mathbb{E}}\left [ \left \| l_{gt} - l_{pred}  \right \|\right ] + \left (  1 - \frac{l_{gt} \cdot l_{pred}}{\| l_{gt} \| \| l_{pred} \|} \right )
    \end{aligned}
\end{equation}
where $l_{pred}$ is the composed latent and $l_{gt}$ is extracted from the pretrained motion encoder. Our total loss can be formulated as:
\begin{equation}
    \begin{aligned}
        \mathcal{L}_{edit} = \mathcal{L} + \mathcal{L}_{latent}
    \end{aligned}
\end{equation}

\noindent
\textbf{Dataset Augmentation} We construct our training data by sampling pairs of a source image and a driving sequence from the same video. For each training pair, we augment the driving frames with both random scaling, color jittor, and piecewise affine transformations following Xportrait2 \cite{xportrait2}. This strategy decouples identity from motion within the driving sequence, compelling the model to obtain identity information exclusively from the source image. To make our model robust to significant \textit{\textbf{scale variations}} between source and drive, we apply random crops to the source image. This technique synthesizes diverse training examples across various scales, thereby enhancing the model's robustness to changes in camera distance.

\subsection{Hierarchical Motion Disentanglement and Composition}
\label{method:composite}

\begin{figure}[t]
    \centering
    \setlength{\belowcaptionskip}{-5pt} 
    \includegraphics[width=\columnwidth]{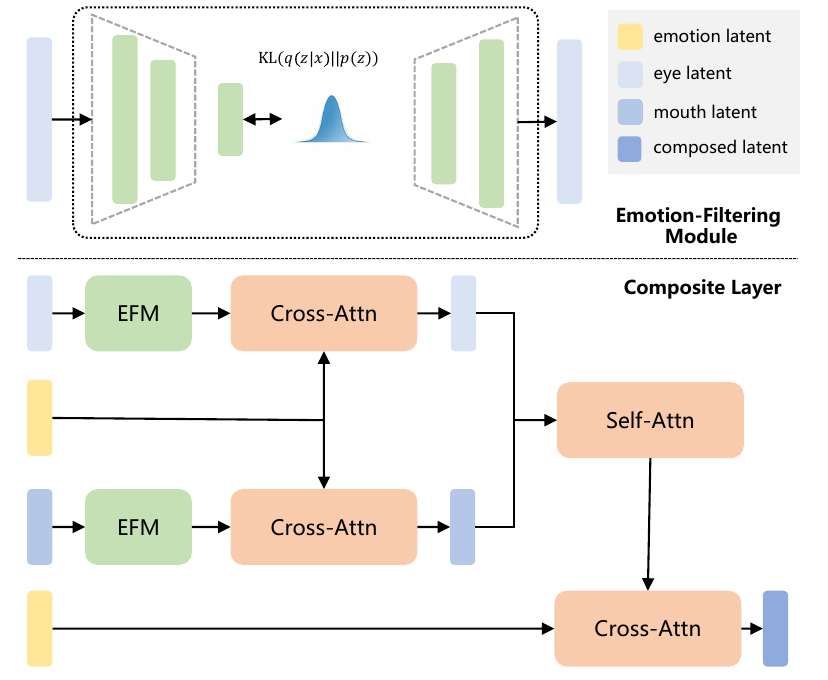}
    \caption{Emotion-Filtering Module (EFM) and Composite Layer. Given the initial motion latents extracted from facial regions, they first pass through the EFM to filter out residual emotional information. Subsequently, these purified latents, along with the target emotion latent, are fed into the composition layer and fused via cross-attention to reconstruct the composed latent.}
    \label{fig:efm}
\end{figure}

A key limitation of existing methods \cite{edtalk,pdfgc} is their reliance on post-hoc decomposition of an entangled motion vector, rather than modeling the generative process. This strategy often results in incomplete separation. The problem is compounded by indirect supervision, which struggles to robustly disentangle distinct facial components.
Built on this insight, our work aims to establish a hierarchical facial model that formulates facial reenactment as a composition process. We explore deconstructing complex motion into a structured representation by designing decomposition strategies tailored to the intrinsic properties of each component. These components are then recomposed to form a composed motion latent that offers an improved balance between high expressiveness and granular control. To this end, 
we introduce a Spatial Layer and a Semantic Layer to decompose the facial motion, and then utilize a Composite Layer to restore a single, highly expressive motion latent.

\noindent
\textbf{Spatial Layer} The objective of this layer is to isolate motion components that are spatially distinct. It decomposes the driving motion by first separating the global head pose from the local facial motion, and then splitting the latter into independent movements. First, to overcome the weak pose control inherent in unified latent space models like XPortrait2~\cite{xportrait2}, we redesign the pose representation to be entirely independent of face expression and introduce a dedicated PoseAdapter to manage its injection, ensuring its injection pathway is completely separate from the expression stream. Specially, We employ MediaPipe \cite{mediapipe} to detect the facial bounding box (bbox), from which we directly derive the head's position $f_w, f_h = x/s, y/s$ and its scale (from the side length) $f_s = \Delta s/s$. The $s$ denotes image size, $x,y$ is the coordinates of box's center point and $\Delta s$ is the maximum dimension of bbox. Subsequently, we employ a pretrained HopeNet \cite{hopenet} model to estimate face rotation angles $f_p,f_y,f_r$. All these pose features are concatenated as $f_{pose}=(f_w,f_h,f_s,f_p,f_y,f_r)$ and processed by PoseEncoder to produce the pose latent $l_{pose}$. This latent is then injected independently into the PoseAdapter to achieve pose control. This structural design grants us explicit, individual control over the head's position, rotation, and scale.

Prior works \cite{edtalk,pdfgc} attempt post-hoc decomposition of a holistic motion latent, but this strategy often fails as the information is already entangled at the encoding stage. We address this by performing separation at the physical level instead. By explicitly segmenting the eye and mouth regions \textit{before} encoding, we ensure a more thorough and robust decoupling of their respective motion latents from the outset. We begin by detecting facial landmarks with MediaPipe \cite{mediapipe} to isolate and crop the eye and mouth regions. These specific regions are then passed to a dedicated MotionEncoder to extract respective eye latent $l_{eye}$ and mouth latent $l_{mouth}$. This input-level separation guarantees that the resulting motion latents are decoupled. Both MotionEncoder share the same architecture and are jointly optimized.

\noindent
\textbf{Semantic Layer} 
While image cropping is effective for separating local facial features, it is ill-suited for disentangling emotion. Emotion is a global, semantic attribute of the entire face, and it is intrinsically coupled with the very movements being isolated. Drawing inspiration from previous methods \cite{edtalk,JiZWWWXC22,pdfgc}, we treat emotion as a low-frequency, temporally smooth component, in contrast to the high-frequency, transient nature of local facial motions. To isolate the emotional latent, we apply a temporal pooling operation over the motion latents extracted from a video window  $l_{emo} = \frac{1}{N} \sum_{i}^{N} l^{i}$, where $l^{i}$ is the motion latent extracted from $i_{th}$ frame. This operation effectively suppresses the high-frequency components corresponding to specific facial movements, yielding a latent representation that encapsulates the persistent emotional state of the segment. For emotion extraction, we employ a MotionEncoder that shares its architecture and pretrained weights with the one used for facial reenactment. It is kept frozen during training, which is forced to rely on its pre-trained knowledge to extract global semantics. This circumvents the risk of local motions leakage such as mouth motion.

\noindent
\textbf{Composite Layer} We assume that facial action is a composite of its fundamental structure and its emotional styling. The former, which we term Emotion-Independent Motion, dictates the core facical motion (e.g., mouth opening). The latter, termed Emotion-Dependent Motion, infuses this base action with expressive nuances (e.g., the tension in the cheeks). To model this interaction, we introduce a Compositional Layer to learn how the global emotional state should modulate the independent eye and mouth movements. The composed latent can be represented as:
\begin{equation}
\label{eq:composite}
\begin{aligned}
    l_{mouth}^{compose} &= l_{mouth}^{basic} + \text{CrossAttn}(l_{mouth}^{basic}, l_{emo}) \\
    l_{eye}^{compose}   &= l_{eye}^{basic} + \text{CrossAttn}(l_{eye}^{basic}, l_{emo}) \\
    l_{full}' &= \text{SelfAttn}(\text{MLP}(l_{eye}^{compose} \oplus l_{mouth}^{compose})) \\
    l_{full}  &= \text{CrossAttn}(l_{full}', l_{emo})
\end{aligned}
\end{equation}

\noindent
where $l_{mouth}^{basic}$ is mouth latent without emotion, $l_{eye}^{basic}$ is eye latent without emotion, $l_{full}$ is the composed latent. The detail is shown in Fig.\ref{fig:efm}. Specifically, we first use cross-attention to inject emotion into the eye latent and mouth latent separately. Second, a self-attention layer fuses these enhanced latents, resolving conflicts and modeling their interplay. Finally, a second cross-attention layer allows the global emotion to modulate the fully fused latent. We will detail the methodology for deriving the Emotion-Independent latents in Sec.\ref{method:EFM}.

\begin{figure*}[htb]
    \centering
    \includegraphics[width=\textwidth]{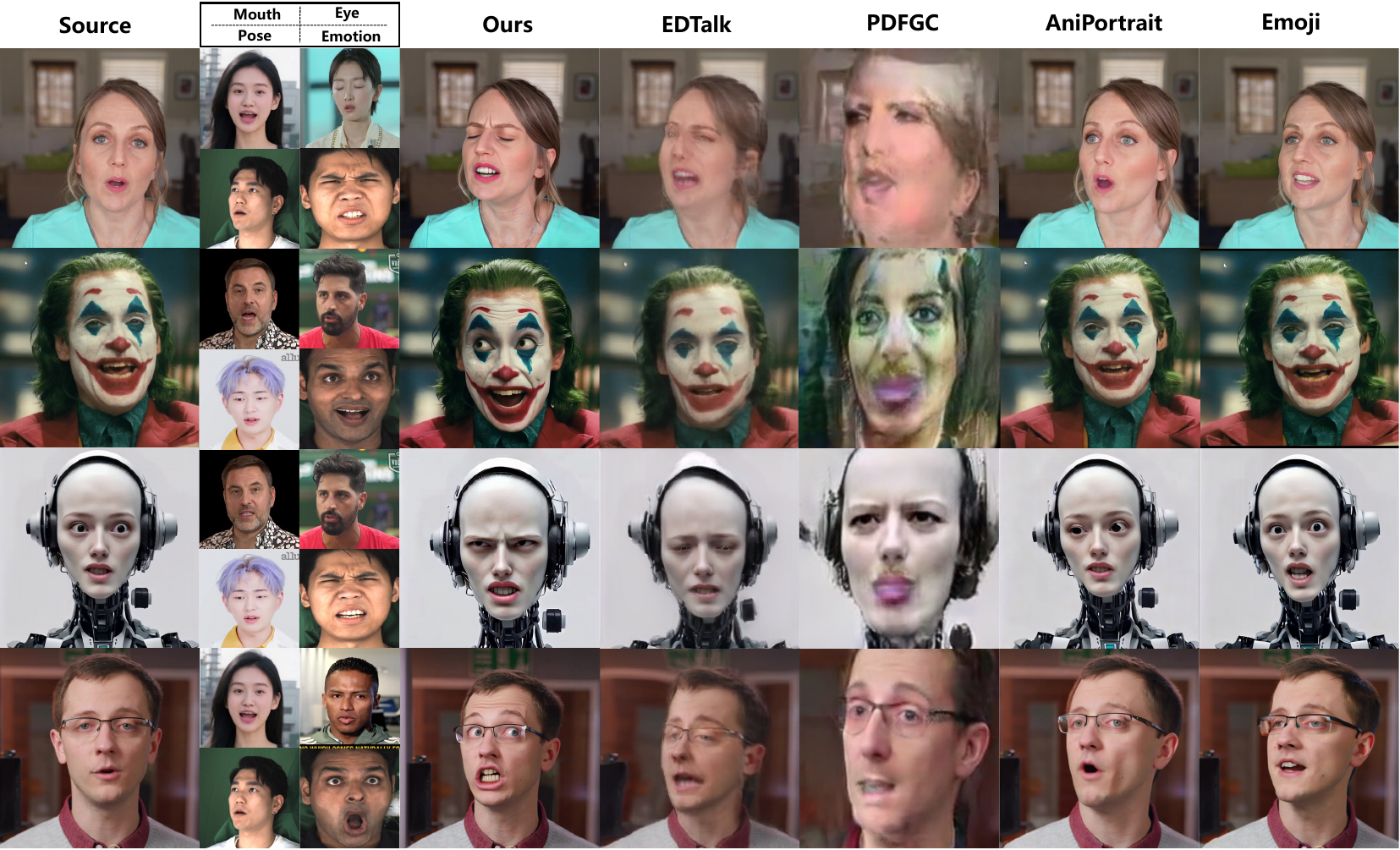}
    \caption{Qualitative comparion. The second column represents the driving signals used to control different facial attributes. AniPortrait \cite{aniportrait} and Emoji \cite{emoji} offer only control over pose and expression, and their expression signal is derived exclusively from the mouth image.}
    \label{fig:comparsion}
\end{figure*}

\begin{table*}[!t]
\centering
\begin{tabular}{c ccc ccc} 
\toprule
\multirow{2}{*}{Method} & \multicolumn{3}{c}{Self-Reenactment} & \multicolumn{3}{c}{Cross-Reenactment} \\ 
\cmidrule(lr){2-4} \cmidrule(lr){5-7}
                        & MSE $\downarrow$        & SSIM $\uparrow$      & LPIPS $\downarrow$     & ID-SIM $\uparrow$   & Pose $\downarrow$     & Expression $\downarrow$   \\ 
\midrule
AniPortrait \cite{aniportrait}             & 0.032       & 0.576      & 0.429      & 0.886     & 8.347     & 0.027         \\
Emoji \cite{emoji}                  & 0.033       & 0.570      & 0.428      & \textbf{0.892}     & 9.543     & 0.026         \\
HunyuanPortrait \cite{hunyuan}         & 0.037       & 0.550      & 0.446      & 0.866     & 19.037    & 0.021         \\
FantasyPortrait \cite{fantasy}         & 0.036       & 0.553      & 0.450      & 0.885     & 19.546    & 0.018         \\
XPortrait2 \cite{xportrait2}              & 0.046       & 0.540      & 0.462      & 0.823     & 4.894     & \textbf{0.010}         \\
EDTalk \cite{edtalk}                  & 0.022       & 0.641      & 0.405      & 0.792     & 8.169     & 0.017         \\
PDFGC \cite{pdfgc}                  & 0.047       & 0.531      & 0.563      & 0.756     & 19.401    & 0.026         \\ 
\midrule
\textbf{Ours}           & \textbf{0.018} & \textbf{0.654} & \textbf{0.316} & 0.880 & \textbf{4.707} & \textbf{0.010} \\ 
\bottomrule
\end{tabular}
\caption{Quantitative comparison of different methods on VFHQ \cite{DBLP:conf/cvpr/XieWZDS22}.
Best results are highlighted in \textbf{bold}.}
\label{tab:quantitative_results_centered}
\end{table*}

\subsection{Emotion-Filtering Module}
\label{method:EFM}
Despite the structural decomposition, a key limitation remains: residual emotion contaminates the local component latents. For instance, a mouth latent extracted from drive image inevitably encodes original emotion, which weakens the control authority of the global emotion latent. Existing methods \cite{edtalk,pdfgc} attempt to mitigate emotion leakage by training on neutral-only datasets. However, this solution is brittle as its dependence on data purity restricts expressive capacity, leading to sub-optimal performance.

Our framework achieves the disentanglement of basic motion and emotion through its analysis-by-synthesis design. Specially, we introduce an autoencoder-style Emotion-Filtering Module based on the information bottleneck principle \cite{bottlneck}. During analysis, this module acts as a low-capacity channel for local latents (e.g., mouth latent), forcing them to shed emotional information while retaining basic facial motions. This yields our Emotion-Independent motions. During synthesis, our Composite Layer re-injects the global emotion to form a complete latent. The entire system is optimized end-to-end to reconstruct the original motion in the form of a holistically expressive motion latent, thereby forcing the bottleneck to learn a meaningful separation. 
We employ an Encoder to first map the original motion latent $l_{mouth}$ into a compact feature space (we use 128), and a Decoder to produce Emotion-Independent latent $l_{mouth}^{basic}$. We constrain the middle feature space to be a low-dimensional latent representation and apply a KL divergence loss to regularize the motion latent space. 

\subsection{Real-time Streaming Generation}
\label{method:fast}
To mitigate the high computational cost of inference and enable practical deployment, we introduce a suite of optimizations built upon the Self-Forcing \cite{selfforcing}. Ultimately, our accelerated framework operates at approximately 20 FPS with a 800ms end-to-end latency for streaming inference, utilizing only a single NVIDIA 5090 GPU.

\noindent
\textbf{DIT Distillation} We apply Distribution Matching Distillation (DMD) \cite{dmd} to distill the model, reducing the sampling steps from 20 to 4 with $\text{CFG}=2$. Consistent with prior work \cite{causvid}, we adapt the bidirectional attention mechanism to a causal one. The attention mechanism is further optimized with a sliding window-like mask \cite{talkingmachines}.
The inference process is further accelerated using a KV cache mechanism, with a minimum chunk size of 2 for efficient streaming. 

\noindent
\textbf{VAE Acceleration} In our 4-step streaming model, the Wan-VAE decoder~\cite{wan2025wan} becomes the primary bottleneck, accounting for ~50\% of the inference latency.
To alleviate this, we design a lightweight VAE decoder by significantly reducing the capacity of the original architecture and retraining it following the protocol in~\cite{wan2025wan}. This streamlined decoder lowers computational complexity while preserving fine visual details with far fewer parameters. Specifically, we reduce the layer width to just $\frac{1}{4}$ of the original Wan-VAE and train it using the following objective:
\begin{equation}
    \mathcal{L}_{vae} = \|x_{pred} - x_{gt}\|^2_2 + \lambda\mathcal{L}_{LPIPS}(x_{pred}, x_{gt})
\end{equation}
where $\lambda=1$ in our experiments. The resulting lightweight VAE achieves a $4\times$ speedup over the original decoder while maintaining comparable reconstruction quality.

%% file: sec/4_experiment.tex
\section{Experiment}


\subsection{Implementation Details}
We train our model on a combined dataset of VFHQ \cite{DBLP:conf/cvpr/XieWZDS22}, NerSemble \cite{nersemble}, and MEAD \cite{mead}. The data is preprocessed by segmenting videos into clips of up to 30 seconds, yielding approximately 300K clips, with all frames resized to $512 \times 512$. The backbone of our model is Wan-I2V \cite{wan2025wan}, initialized with its pretrained weights. The training follows a two-stage procedure. First, the model is trained for the face reenactment task for 100K steps on 24 A100 GPUs. Afterward, we freeze the DiT backbone, PoseEncoder and MotionEncoder for emotion, proceed with a finetuning stage for facial disentanglement. This second stage runs for 20K steps on 24 A100 GPUs. For detailed training configurations, please refer to the supplementary material.

\begin{table}[t]
\centering
\setlength{\tabcolsep}{4pt} 
\begin{tabular}{l
                S[table-format=2.3] 
                S[table-format=1.3] 
                S[table-format=1.3] 
                S[table-format=1.3] 
                }
\toprule
Method      & {Pose $\downarrow$} & {Expression $\downarrow$} & {Mouth $\downarrow$} & {Eye $\downarrow$} \\ 
\midrule
EDTalk \cite{edtalk}     & 25.999 & 0.028      & 0.014 & {--}    \\ 
PDFGC \cite{pdfgc}       & 33.081 & 0.035      & 0.025 & 0.046 \\
AniPortrait \cite{aniportrait} & 19.200 & 0.032      & {--}    & {--}    \\
Emoji \cite{emoji}       & 16.160 & 0.029      & {--}    & {--}    \\
\midrule
\textbf{Ours} & \textbf{13.695} & \textbf{0.023} & \textbf{0.012} & \textbf{0.033} \\ 
\bottomrule
\end{tabular}
\caption{Quantitative Comparison for facial components control accuracy using MEAD \cite{mead} dataset. \textbf{Bold} indicates the best result.}
\label{tab:control}
\end{table}

\subsection{Comparison with baselines}

\subsubsection{Qualitative results}
The qualitative comparison results for fine-grained facial control are shown in Fig.\ref{fig:comparsion}. We observe that methods such as EDTalk \cite{edtalk} and PDFGC \cite{pdfgc} tend to produce results with visual artifacts and exhibit a notable sensitivity to scale variations, which hinders fine-grained control over individual components. While Follow-Your-Emoji \cite{emoji}(abbreviated as Emoji for brevity) and AniPortrait \cite{aniportrait} achieve coarse-grained pose-expression decoupling, they are not designed for fine-grained component manipulation. Moreover, their outputs can be undesirably influenced by the reference image's inherent expression, limiting their editing flexibility. In contrast,
our method achieves more effective and fine-grained control over individual facial motion components, while also exhibiting enhanced robustness to pose and scale variations, greater flexibility in generating target expressions, and strong generalization across diverse visual styles. For comparisons with face reenactment methods, please refer to the supplementary material. 

\subsubsection{Quantitative results}

For quantitative evaluation, we assess our method on facial reenactment and facial components control tasks. For facial reenactment, we sample 200 videos (48 frames each) from VFHQ \cite{DBLP:conf/cvpr/XieWZDS22} as targets and 200 different images for cross-identity references. Self-Reenactment quality is measured by MSE, SSIM, and LPIPS, while Cross-Reenactment performance is evaluated using identity, pose, and expression similarity. To assess the control precision over individual facial components, we design an experiment using 40 videos and 40 images from MEAD~\cite{mead}. In this setup, we drive only a single component at a time (e.g., eyes or mouth) and report the similarity in the corresponding region to quantify control accuracy. We randomly scale the reference images to simulate image \textit{\textbf{scale variations}}. For metrics, identity similarity (ID-SIM) is the cosine distance between ArcFace \cite{arcface} features. Pose and expression similarities are assessed by the L2 distances between facial params extracted via pretrained estimators \cite{hopenet, mediapipe}. 

As shown in Tab.\ref{tab:quantitative_results_centered} and Tab.\ref{tab:control}, our method demonstrates competitive performance, surpassing most competing approaches across a range of tasks. This advantage stems from our model's fine-grained decomposition and faithful reconstruction of facial components. A notable advantage of our method is its improved robustness to scale variations. 

Many previous works suffer from scale entanglement. For instance, XPortrait2 \cite{xportrait2} achieves high expression accuracy at the expense of higher reconstruction loss by rigidly adhering to the reference image's scale. In contrast, our explicit decomposition of pose mitigates this limitation.
Furthermore, Tab.\ref{tab:control} highlights our model's capability in facial component control. Compared with existing approaches, our method achieves more precise control over a wider array of individual facial components.

\subsection{Ablation study}

\begin{figure}[t]
    \centering
    \includegraphics[width=\columnwidth]{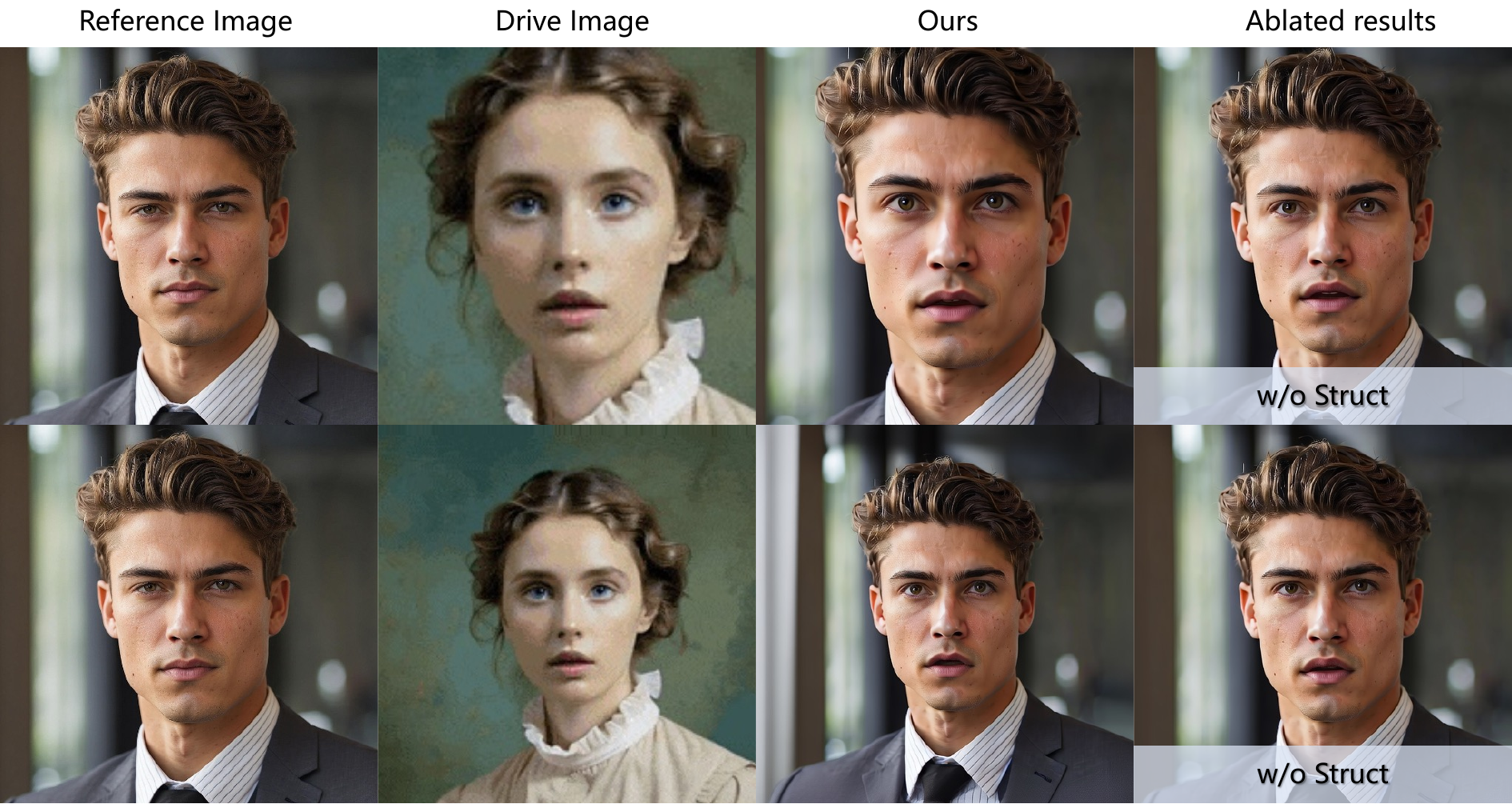}
    \caption{Ablation study of pose-expression disentanglement.}
    \label{fig:ablation_pose}
\end{figure}

\begin{figure}[t]
    \centering
    \includegraphics[width=\columnwidth]{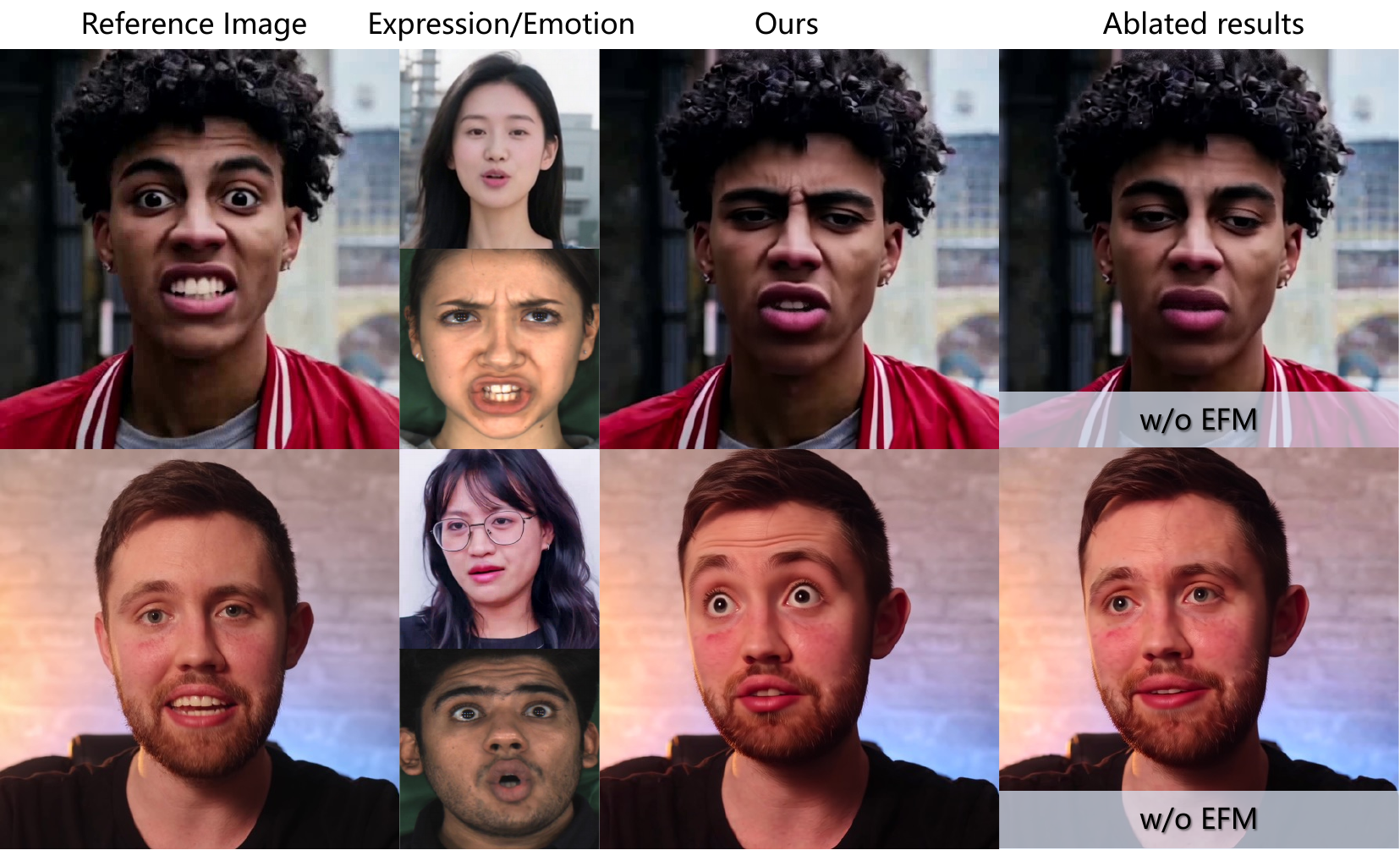}
    \caption{Ablation study of expression-emotion disentanglement.}
    \label{fig:ablation_efm}
\end{figure}

We conduct ablation studies to analyze the impact of our key contributions: the effect of the proposed structural decomposition on pose-expression disentanglement, and the role of the Emotion-Filtering Module (EFM) in the decomposition and recombination of the emotion component.

\noindent
\textbf{Pose-Expression Disentanglement} We validate our structural decomposition by creating a variant ("w/o Struct") inspired by Xportrait2 \cite{xportrait2}, which discards our PoseAdapter and instead fuses pose and expression latents via an MLP into a unified motion latent. This approach exhibits severe scale entanglement, where the model inherits the scale from the reference image instead of the driving pose, as shown in Fig.\ref{fig:ablation_pose}. This issue, also prevalent in Xportrait2, validates the critical role of our explicit pose-expression separation in achieving robust disentanglement. Additional disentanglement results are provided in the supplementary material.

\noindent
\textbf{Expression-Emotion Disentanglement} We verify the EFM by replacing it with a parameter-matched MLP while removing its bottleneck design and KL loss (the "w/o EFM" variant). As shown in Fig.\ref{fig:ablation_efm}, removing the EFM has a noticeable impact. The model without EFM exhibits a reduced capacity to suppress the source expression, resulting in a diminished ability for independent expression control. Conversely, our full model leverages the EFM to effectively isolate basic facial motion from the source's emotion, thus achieving robust control over the final expression.

%% file: sec/5_conclusion.tex
\section{Conclusion}
In this paper, we introduced PortraitDirector, a novel framework for high-fidelity and controllable facial reenactment. 
We explore formulating  face reenactment as a composition task. Our framework achieves control by routing different facial components to dedicated layers: spatial, semantic, and composite. Each layer is specifically designed to handle a distinct form of motion. 
Furthermore, we presented an Emotion-Filtering Module that effectively decouples facial motion from emotional expression, mitigating the issue of source-emotion bias. Finally, through a series of dedicated optimizations, including DMD and causal attention mechanism, PortraitDirector achieves real-time, streaming performance on a single consumer GPU. Our method delivers superior performance across diverse scenes, enabling real-time and highly controllable facial animation.